\documentclass{article}
\usepackage{ijcai23}

\usepackage{times}
\usepackage{soul}
\usepackage{url}
\usepackage[hidelinks]{hyperref}
\usepackage[utf8]{inputenc}
\usepackage[small]{caption}
\usepackage{graphicx}
\usepackage{amsmath}
\usepackage{amsthm}
\usepackage{booktabs}
\usepackage[switch]{lineno}

\usepackage{subfigure} 
\usepackage{amssymb}
\usepackage{multirow}
\usepackage{multicol}

\usepackage{algorithm}  
\usepackage[noend]{algpseudocode}

\newcommand{\tableref}[1]{Table~\ref{#1}} 
\newcommand{\figref}[1]{Figure~\ref{#1}} 

\graphicspath{{image/}}
 \usepackage{enumitem}
\usepackage{fancyhdr}
\usepackage{adjustbox}

\title{Causal-Based Supervision of Attention in Graph Neural Network:\\
A Better and Simpler Choice towards Powerful Attention}

 \author{
Hongjun Wang \textsuperscript{\rm1}
\and
Jiyuan Chen \textsuperscript{\rm1} \and
Lun Du \textsuperscript{\rm2} \and
Qiang Fu \textsuperscript{\rm2} \and
Shi Han \textsuperscript{\rm2} \and
Xuan Song \textsuperscript{\rm1}
\affiliations
$^1$Southern University of Science and Technology, Shenzhen, China\\
$^2$Microsoft Research Asia, Beijing, China\\
\emails
\{wanghj2020, chenjy6, songx\}@mail.sustech.edu.cn,\\
\{lun.du, qifu,  shihan\}@microsoft.com
}

\begin{document}

\maketitle

\begin{abstract}
Recent years have witnessed the great potential of attention mechanism in graph representation learning. However, while variants of attention-based GNNs are setting new benchmarks for numerous real-world datasets, recent works have pointed out that their induced attentions are less robust and generalizable against noisy graphs due to lack of direct supervision. In this paper, we present a new framework which utilizes the tool of causality to provide a powerful supervision signal for the learning process of attention functions. Specifically, we estimate the direct causal effect of attention to the final prediction, and then maximize such effect to guide attention attending to more meaningful neighbors. Our method can serve as a plug-and-play module for any canonical attention-based GNNs in an end-to-end fashion. Extensive experiments on a wide range of benchmark datasets illustrated that, by directly supervising attention functions, the model is able to converge faster with a clearer decision boundary, and thus yields better performances.
\end{abstract}

 \section{Introduction}
 Graph-structured data is widely used in real-world domains, such as social networks \cite{zhang2018link}, recommender systems \cite{wu2022graph}, and biological molecules
 \cite{gilmer2017neural}. The non-euclidean nature of graphs has inspired a new type of machine learning model, Graph Neural Networks (GNNs) \cite{kipf2016semi,defferrard2016convolutional,du2022gbk}.
Generally, GNN iteratively updates features of the center node by aggregating those of its neighbors and has achieved remarkable success across various graph
analytical tasks. However, the aggregation of features between unrelated nodes has long been an obstacle for GNN, keeping it from further improvement.

Recently, Graph Attention Network (GAT) \cite{velivckovic2017graph} pioneered the adoption of the attention mechanism, a well-established method with proven effectiveness in deep learning \cite{vaswani2017attention}, into the neighborhood aggregation process of GNNs to alleviate the issue. The key concept behind GAT is to adaptively assign importance to each neighbor during the aggregation process. Its simplicity and effectiveness have made it the most widely used variant of GNN.  Following this line, a myriad of attention-based GNNs have been proposed and have achieved state-of-the-art performance in various tasks \cite{sun2021scalable,zhang2022graph,ying2021transformers,brody2021attentive}.





Nevertheless, despite the widespread use and satisfying results, in the past several years, researchers  began to rethink if the learned attention functions are truly effective \cite{kim2022find,wang2019improving,wu2022causally,knyazev2019understanding,liu2021non,wang2021evolving}. 
As we know, most existing attention-based GNNs learn the attention function in a weakly-supervised manner,
where the attention modules are simply supervised by the final loss function, without a powerful supervising signal to
guide the training process. And the lack of direct supervision on attention might be a potential cause of a less robust and 
generalizable attention function against real-world noisy graphs \cite{sui2022causal,knyazev2019understanding,li2022finding,wang2019improving,ye2021sparse,huang2023robust}. To address this problem, existing work enhances the quality of attention through auxiliary regularization terms (supervision). However, concerns have been raised that these methods often rely heavily on human-specified prior assumptions about a specific task, which limits their generalizability \cite{you2020does,wu2022causally}. Additionally, the auxiliary regularization is formulated independently of the primary prediction task, which may disrupt the original optimization target and cause the model to ``switch" to a different objective function during training \cite{kim2022find,you2020does}. 

\begin{figure}
    \centering
    \includegraphics[width=0.9\linewidth]{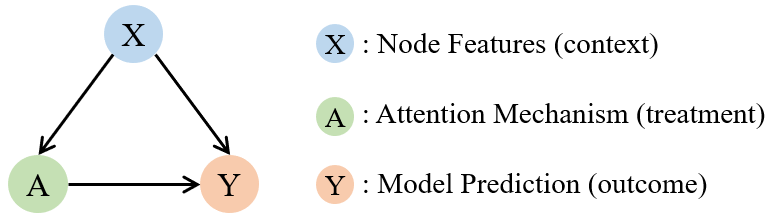}
	\caption{ Structural causal model of attention-based GNNs}
    \label{fig:casualscm}
\end{figure}

Recently, causal inference \cite{pearl2009causality} has attracted many researchers in the field of GNNs by utilizing structural causal model (SCM) \cite{cinelli2019sensitivity} to handle distribution shift \cite{zhao2022learning} and shortcut learning \cite{feng2021should}. In this paper, we argue that the tool of causal inference has also shed light on a promising avenue that could supervise and improve the quality of GNN's attention directly, while in the meantime we will not make any assumptions about specific tasks or models, and the supervision signal for attention implicitly aligns well with the primary task. Before going any deeper, we first provide a general schema for the SCM of attention-based GNNs in Figure \ref{fig:casualscm}, which uses nodes to represent variables and edges to indicate causal relations between variables. As we can see, after a high-level abstraction, there are only three key factors in SCM, including the node features $X$, attention maps $A$, and the model's final prediction $Y$. Note that in causal language, $X$, $A$, and $Y$ also denotes the context, treatment, and outcome respectively. For edges in SCM, the link $X \rightarrow A$ represents that the attention generation relies on the node's features (i.e., context decides treatment). And links $(X,A) \rightarrow Y$ indicate that the model's final prediction is based on both the node's features $X$ and the attention $A$ (i.e., the final outcome is jointly determined by both context and treatment). 



In order to provide a direct supervision signal and further enhance the learning of attention functions, the first step would be finding a way to measure the quality of attention (i.e., quantifying  what to improve). Since there are no unified criteria on the way of measurement, researchers usually propose their own solution according to the tasks they are facing, and this is very likely to introduce the unfavorable human-intervened prior assumptions \cite{you2020does}. For example, CGAT \cite{wang2019improving} believes that better attention should focus more on one-hop neighbors. While this assumption surely works on homophilous graphs, it will suffer from huge performance degradation in heterophilous scenarios. Our method differs a lot from existing work in that we introduce SCM to effectively decouple the direct causal effect of attention on the final prediction (i.e., link $A \rightarrow Y$), and use such causal effect as a measurement for the quality of attention. In this way, it is the model and data that decide if the attention works well during training instead of human-predefined rules. And this has been shown to be non-trivial in various machine learning fields because what might seem reasonable to a human might not be considered the same way by the model \cite{kumar2010self,wang2022easy}. Another drawback of existing attention regularization methods, as previously mentioned, is the deviation from primary tasks. SuperGAT \cite{kim2022find} uses link prediction to improve the attention quality for node classification, but as the author claims in the paper, there is an obvious trade-off between the two tasks. In this paper, we alleviate this problem by directly maximizing the causal effect of attention on the primary task (i.e., strengthening the causal relation $A \rightarrow Y$). Under mild conditions, we can deem the overall optimization is still towards the primary objective, except that we additionally provide a direct and powerful signal for the learning of attention in a fully-supervised manner.

In summary, this paper presents a \textbf{C}ausal \textbf{S}upervision for \textbf{A}ttention in graph neural networks (abbreviated as \textbf{CSA} in the following paragraphs). CSA has strong applicability because no human-intervened assumptions are made on the target models or training tasks. And the supervision of CSA can be easily and smoothly integrated into optimizing the primary task to performing end-to-end training. We list the main contributions in this paper as follows:

\begin{itemize}[leftmargin=*]
\item[$\bullet$] We explore and provide a brand-new perspective  to directly boost GNN's attention with the tool of causality. To the best of our knowledge, this is a promising direction that still remains unexplored.
\item[$\bullet$] We propose CSA, a novel causal-based supervision framework for attention in GNNs, which can be formulated as a simple yet effective external plug-in for a wide range of models and tasks to improve their attention quality.  
\item[$\bullet$] We perform extensive experiments and analysis on CSA and the universal performance gain on standard benchmark datasets validates the effectiveness of our
design.

\end{itemize}

 \section{Related Work}
 \noindent \textbf{Attention-based Graph Neural  Networks.}
Modeling pairwise importance between elements in graph-structured data dates back to interaction networks \cite{battaglia2016interaction,hoshen2017vain} and relational networks \cite{santoro2017simple}.
Recently GAT \cite{velivckovic2017graph} rose as one of the representative work of  attention-based GNNs using self-attention \cite{vaswani2017attention}. The remarkable success of GAT in multiple tasks has motivated many works focusing on integrating attention into GNN \cite{thekumparampil2018attention,zhang2018gaan,wang2022st,zhang2020adaptive,gao2019graph,hou2022measuring}. \textit{Lee et al.} have also conducted a comprehensive survey
\cite{lee2018attention} on various types of attention used in GNNs.


\noindent \textbf{Causal Inference in Graph Neural Network.} Causality  \cite{pearl2014interpretation} provides researchers
new methodologies to design robust measurements,
discover hidden causal structures and confront data biases. A myriad of studies has shown that incorporating causality is beneficial to graph neural network in various tasks. \cite{zhao2022learning} makes use of counterfactual links to augment data for link prediction improvement. \cite{sui2022causal} performs interventions on the representations of graph data to identify the causally attended
subgraph for graph classification. \cite{feng2021should} on the other hand, applies causality to estimate the causal effect of node's local structure to assist node classification.


\noindent \textbf{Improving Attention in GAT.}
There is a great number of work dedicated to improving attention learning in GAT. \cite{kim2020find} enhances attention by exploiting two attention forms compatible with a self-supervised task to predict edges. \cite{brody2021attentive} introduces a simple fix by modifying the order of operations in GAT.
\cite{wang2019improving} develops an approach using constraint on the attention weights according to the class boundary and feature aggregation pattern. In addition, causality also plays a role in boosting the attention of GATs recently. \cite{wu2022causally} estimates the causal effect of edges by intervention and regularizes edges' attention weights according to their causal effects. 

\section{Preliminaries}\label{sec:prelim}
We start by introducing the notations and formulations of graph neural networks and their attention variant.  
Let $\mathcal{G}=\{\mathcal{V},\mathcal{E}$\} represents a graph where $\mathcal{V}=\{v_i\}_{i=0}^{n}$ is the set of nodes and $\mathcal{E} \in \mathcal{V} \times \mathcal{V}$ is the set of edges. 
For each node $v \in \mathcal{V}$, it has its own neighbor set $N(v)=\{ u \in \mathcal{V} \mid (v,u) \in  \mathcal{E}\}$ and its initial feature vector $x_v^0 \in \mathbb{R}^{d^{0}}$, where $d^0$ is the original feature dimension. 
Generally, GNN follows the message-passing mechanism to perform feature updating, where each node's feature representation is updated by aggregating the representations of its neighbors and then combining the aggregated messages
with its ego representation \cite{xu2018powerful}.
Let $m_v^l \in \mathbb{R}^{d^{l}}$ and  $x_{v}^{l} \in \mathbb{R}^{d^{l}}$ be the message vector and representation vector of node $v$ at layer $l$, we formally define the updating process of GNN as:
\begin{align*}\label{eq:aggcom}
m_v^l&=\operatorname{AGGREGATE}\left(\left\{x_{j}^{(l-1)}, \forall j \in N(v)\right\}\right)\\
x_v^l&=\operatorname{COMBINE}\left(x_v^{l-1}, m_v^l\right),
\end{align*}
where $\operatorname{AGGREGATE}$ and $\operatorname{COMBINE}$ are aggregation functions (e.g., mean, LSTM) and combination function (e.g., concatenation), respectively. 
The design of these two functions is what mostly distinguishes one type of GNN from the other.
GAT \cite{velivckovic2017graph}
augments the normal aggregation with the introduction of self-attention.
The core idea of self-attention in GAT is to learn a scoring function that computes an attention score for every node in $N(v)$ to indicate their relational importance to node $v$. In layer $l$, such process is defined by the following equation:
\begin{equation*}
e\left(x_{v_i}^{l}, x_{v_j}^{l}\right)=
	\sigma
	\left(
		(\mathbf{a}^{l})^{\top}\cdot\left[ W^{l} \ x_{v_i}^{l} \| W^{l} \ x_{v_j}^{l} \right]
	\right),
	\label{eq:gat}
\end{equation*}
where ($\mathbf{a}^{l}, W^{l}$), $\sigma$ are learnable matrices and activation function (e.g., LeakyReLU) respectively, and $\|$ denotes vector concatenation. 
The attention scores are then normalized across all neighbors $v_j\in N(v_i)$ using softmax to ensure consistency:
\begin{equation*}
	\alpha_{ij}^{l} =
\frac{\mathrm{exp}\left(e\left(x_{v_i}^{l} , x_{v_j}^{l}\right)\right)}{\sum\nolimits_{v_j\in {N}(v_i)} \mathrm{exp}\left(e\left(x_{v_i}^{l} , x_{v_j}^{l}\right)\right)}
	\label{eq:softmax}
\end{equation*}
Finally, GAT computes a weighted average of the features of the neighboring nodes as the new feature of $v_i$, which is demonstrated as follows:
\begin{equation*}\label{eqn:gat}
	x_{v_i}^{l+1}=
\sigma\left(\sum\nolimits_{v_j\in {N}(v_i)}
		\alpha_{ij}^{l}
		 W^{l} \ x_{v_j}^{l}
	\right).
\end{equation*}
 \begin{figure}
	\centering
	\includegraphics[width=1\linewidth]{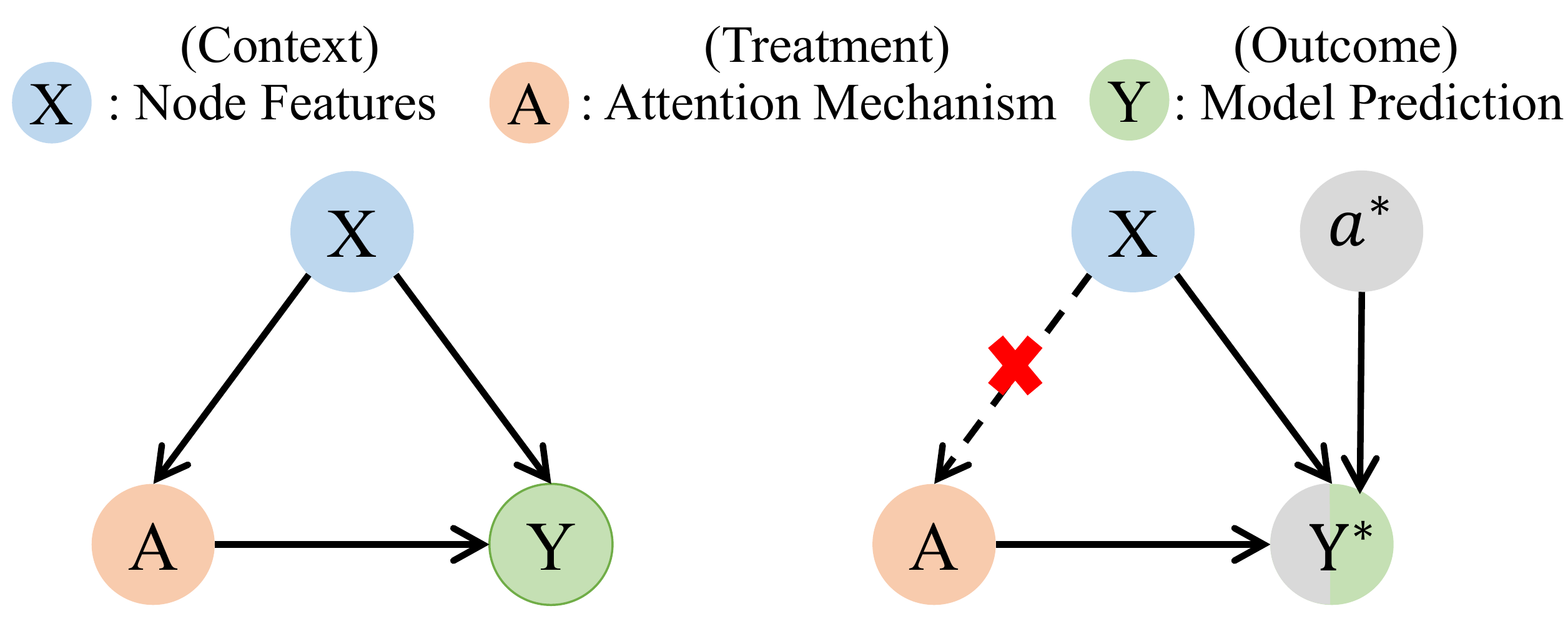}
	\caption{Deriving causal effects through counterfactual}\label{fig:causal}
\end{figure}


\section{Casual-based Supervision on GNN's Attention}

In this section, we first introduce how the causal effect of attention can be derived from the structural causal model of attention-based GNNs. Specifically, this is done with the help of the widely used counterfactual analysis in causal reasoning. After that, with the obtained causal effects, we elaborate three candidate schemes to incorporate with the training of attention-based GNNs to improve their quality of attention.


\subsection{Causal Effects of Attention}
As previously mentioned, the first step towards improving attention lies in measuring the quality of existing attention. However, since deep learning models usually exhibit as black boxes, it is generally infeasible to directly assess their attention qualities. Existing works mainly address this issue by introducing human priors to build pre-defined rules for some specific models and tasks. Yet, it has been a long debate on whether human-made rules share consensus with deep learning models during training \cite{kumar2010self,ribeiro2016should}. Fortunately, the recent rise of causal inference technology has offered effective tools 
to help us think beyond the black box and analyze causalities between model variables, which leads us to an alternative way to directly utilize the causal effect of attention to measure its quality. Since the obtained causal effects are mainly affected by the model itself, it is a more accurate and unbiased measurement of how well the attention actually learns.  

We first give a brief review of the formulation of attention-based graph neural network in causal languages, as shown in \figref{fig:causal}(a). The generated attention map $A$ is directly affected by node feature $X$. And the model prediction $Y$ is jointly determined by both $X$ and $A$. We denote the inferring process of the model as:
\begin{equation}\label{eq:cfnot}
Y_{x,a} = Y(X=x,A=a),
\end{equation}
which indicates that model will give value $Y_{x,a}$ if the value of $X$ and $A$ are set to $x$ and $a$ respectively. In order to pursue the attention's causal effect, we introduce the widely-used counterfactual analysis \cite{pearl2022direct} in causal reasoning. 

The core idea of counterfactual causality lies in asking: given a certain data context (node feature $X$),
what the outcome (model prediction $Y$) would have been if the treatment (attention map $A$) had not been the observed value? To answer the imaginary question, we have to manipulate the values of several variables to see the effect, and this is formally termed as \emph{intervention} in causal inference literature, which can be denoted as $do(\cdot)$. In $do(\cdot)$ operation, we compulsively select a counterfactual value to replace the original factual value of the
intervened variable. And once a variable is intervened, its all incoming links in the SCM will be cut off and its value is independently given, while
other variables that are not affected still maintain the original value. In our case, for example, $do(A=a^*)$ means we demand the attention $A$ to take the non-real value $a^*$ (e.g., reverse/random attention) so that the link $X \rightarrow A$ is 
cut-off and $A$ is no longer be
affected by its causal parent $X$. This process is illustrated in \figref{fig:causal}(b) and the mathematical formulation is given as:
\begin{equation}\label{eq:cfnot}
Y_{x,a^*} = Y(X=x,do(A=a^*)),
\end{equation}
which indicates that after $do(\cdot)$ operation which changes the value of attention to be $a^*$, the output value of the model also changes to $Y_{x,a^*}$. Finally, let us consider a case where we assign a \emph{dummy} value $\Tilde{a}$ to the attention map so that for each ego node, all its neighbors share the same attention weights, the feature aggregation of the graph attention model will then degrade to an unweighted average. In this case, according to the theory of causal inferences \cite{vanderweele2016explanation}, the Total Direct Effect (TDE) of attention to model prediction can be obtained by computing the differences of model outcome $Y_{x,a}$ and $Y_{x,\Tilde{a}}$, which is formulated as follows:
\begin{equation}\label{eq:vqande}
    TDE=Y_{x,a}-Y_{x,\Tilde{a}}.
\end{equation}

It is worth noting that the induction of attention's causal effect does not exert any assumptions and constraints, which lays a solid foundation for our wide applicability on any graph attention models.


\begin{figure}
	\centering
	\includegraphics[width=1\linewidth]{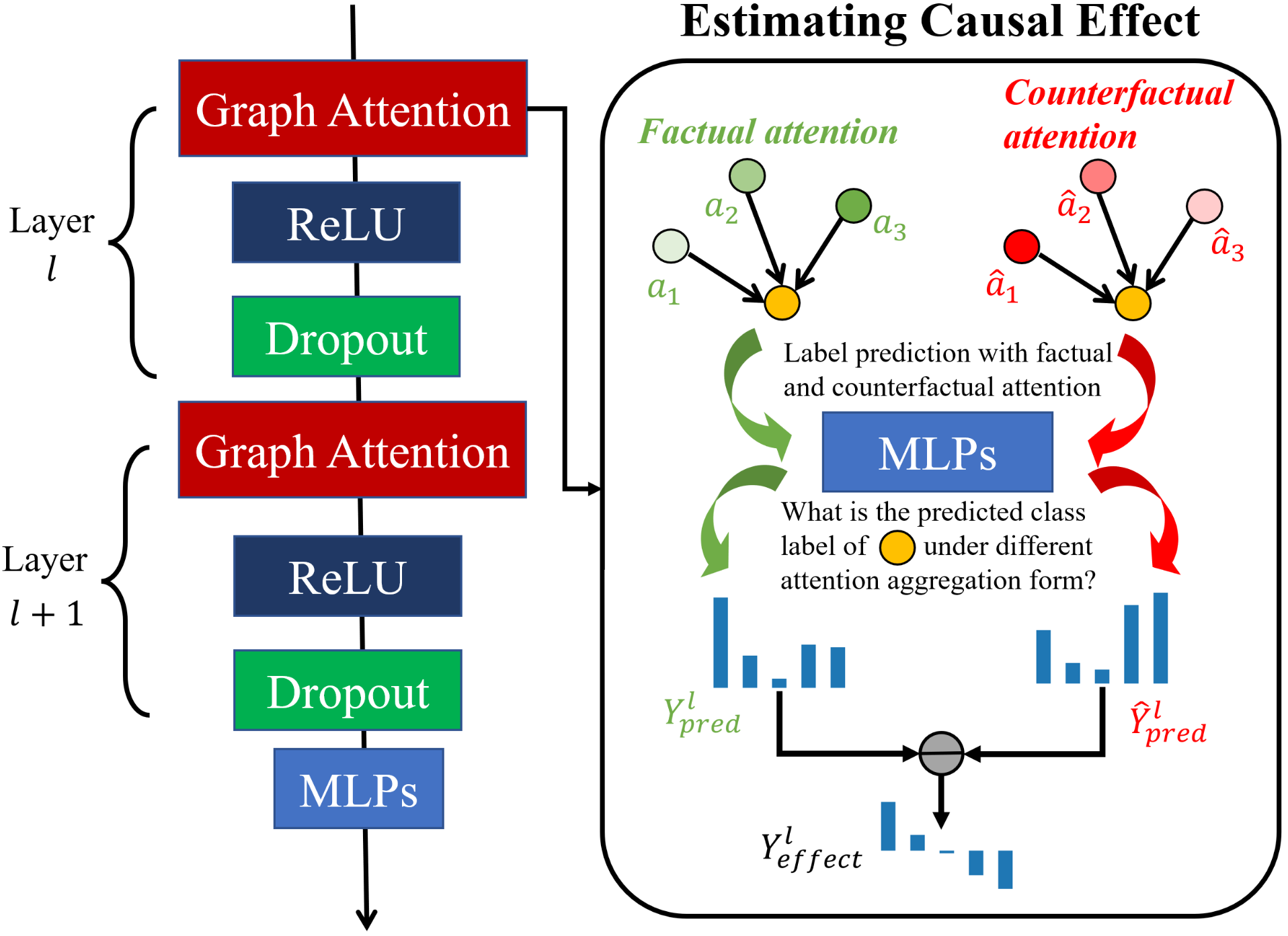}
	\caption{The schematic of CSA is shown above as a plug-in to graph attention methods. The $a$ and $\hat{a}$ indicate the factual and counterfactual attention values, respectively. We subtract the counterfactual classification results from the original classification to
    analyze the causal effects of learned attention (i.e., attention quality) and directly maximize them in the training process towards primary task.
}
	\label{fig:model}
\end{figure}

\subsection{Supervision on Attention with Causal Effects}
\label{learning}


We have already demonstrated the derivation of attention's causal effect in the previous section. In this part, we will discuss how to utilize the obtained causal effect for attention quality improvement.
Previous works that make use of  an auxiliary task to regularize attention usually suffered from performance trade-off between the primary task and auxiliary task. In this work, we alleviate this problem
by directly maximizing the causal effect of attention on the
primary task. The overall schema of this part is shown in \figref{fig:model}.

\begin{table*}[t]
	\centering

	\begin{adjustbox}{width=0.95\textwidth}
		\begin{tabular}{lccccccc}
			\toprule & \textbf{Texas} & \textbf{Wisconsin} & \textbf{Actor} & \textbf{Squirrel} & \textbf{Chameleon} & \textbf{Cornell} & \textbf{Crocodile} \\
			$\mathcal{H}(\mathcal{G})$ & $\mathbf{0 . 1 1}$ & $\mathbf{0 . 2 1}$ & $\mathbf{0 . 2 2}$ & $\mathbf{0 . 2 2}$ & $\mathbf{0 . 2 3}$ & $\mathbf{0 . 3}$ & $\mathbf{0.26}$ \\
			
			\textbf{\#Nodes} & 183 & 251 & 7,600 & 5,201 & 2,277 & 183 & 11,631 \\
			\textbf{\#Edges} & 295 & 466 & 191,506 & 198,493 & 31,421 & 280 & 899,756 \\
			\textbf{\#Classes} & 5 & 5 & 5 & 5 & 5 & 5 &  6 \\
			\textbf{\#Features} & 1703 &1703& 932 &2089&2325&1703&500\\ 
			\hline

        MLP 	& 81.32 $\pm$ 6.22 & 84.38 $\pm$ 5.34 & 36.09 $\pm$ 1.35 & 28.98 $\pm$ 1.32 & 46.21 $\pm$ 2.89 & 83.92 $\pm$ 5.88 & 54.35 $\pm$ 1.90 \\
        SGCN & 56.41 $\pm$ 4.29 & 54.82 $\pm$ 3.63 & 30.50 $\pm$ 0.94 & 52.74 $\pm$ 1.58 & 60.89 $\pm$ 2.21 & 62.52 $\pm$ 5.10 & 51.80 $\pm$ 1.53 \\
        GCN 	& 55.59 $\pm$ 5.96 & 53.48 $\pm$ 4.75 & 28.40 $\pm$ 0.88 & 53.98 $\pm$ 1.53 & 61.54 $\pm$ 2.59 & 60.01 $\pm$ 5.67 & 52.24 $\pm$ 2.54 \\
        H2GCN 	& 84.81 $\pm$ 6.94 & 86.64 $\pm$ 4.63 & 35.83 $\pm$ 0.96 & 37.95 $\pm$ 1.89 & 58.27 $\pm$ 2.63 & 82.08 $\pm$ 4.71 & 53.10 $\pm$ 1.23 \\
        APPNP 	& 81.93 $\pm$ 5.77 & 85.48 $\pm$ 4.58 & 35.90 $\pm$ 0.96 & 39.08 $\pm$ 1.76 & 57.80 $\pm$ 2.47 & 81.92 $\pm$ 6.12 & 53.06 $\pm$ 1.90 \\
        GPR-GNN & 79.44 $\pm$ 5.17 & 84.46 $\pm$ 6.36 & 35.11 $\pm$ 0.82 & 32.33 $\pm$ 2.42 & 46.76 $\pm$ 2.10 & 79.91 $\pm$ 6.60 & 52.74 $\pm$ 1.88 \\
        
        \hline
        GAT& 55.21 $\pm$ 5.70 & 52.80 $\pm$ 6.11 & 29.04 $\pm$ 0.66 & 40.00 $\pm$ 0.99 & 59.32 $\pm$ 1.54 & 61.89 $\pm$ 6.08 & 51.28 $\pm$ 1.79 \\
        +CSA-I & 56.17 $\pm$ 5.32 & 53.23 $\pm$ 6.28 & 29.03 $\pm$ 0.79 & 40.51 $\pm$ 0.98 & 60.73 $\pm$ 1.35 & 62.75 $\pm$ 6.32 & 51.67 $\pm$ 1.62  \\
        +CSA-II & \bfseries 58.21 $\pm$ 4.79 & \bfseries 54.35 $\pm$ 6.54 & 29.71 $\pm$ 0.74 & 41.02 $\pm$ 1.23 & \bfseries 61.31 $\pm$ 1.13 &\bfseries 64.26 $\pm$ 5.21 & \bfseries 52.20 $\pm$ 1.74 \\
        +CSA-III & 58.04 $\pm$ 5.27 & 53.98 $\pm$ 6.30 & \bfseries 29.72 $\pm$ 0.86 & \bfseries 41.38 $\pm$ 1.19 &  61.20 $\pm$ 1.37 & 63.58 $\pm$ 6.03 & 52.13 $\pm$ 1.83 \\
        \hline	 
        FAGCN 	& 82.54 $\pm$ 6.89 & 82.84 $\pm$ 7.95 & 34.85 $\pm$ 1.24 & 42.55 $\pm$ 0.86 & 61.21 $\pm$ 3.13 & 79.24 $\pm$ 9.92 & 54.35 $\pm$ 1.11 \\
        +CSA-I & 82.65 $\pm$ 7.11 & 83.37 $\pm$ 7.79 & 34.77 $\pm$ 0.95 & 42.55 $\pm$ 0.74 & 61.86 $\pm$ 2.98 & 80.01 $\pm$ 9.72 & 54.44 $\pm$ 1.18 \\
        +CSA-II & 83.29 $\pm$ 6.80 & 83.11 $\pm$ 8.26 & 34.88 $\pm$ 0.86 & 42.58 $\pm$ 0.93 & 61.74 $\pm$ 3.39 & \bfseries 81.35 $\pm$ 9.68 & 54.45 $\pm$ 1.23 \\
        +CSA-III & \bfseries 84.72 $\pm$ 6.71 & \bfseries 84.23 $\pm$ 7.21 & \bfseries 35.12 $\pm$ 0.98 & \bfseries 43.38 $\pm$ 1.02 & \bfseries 62.52 $\pm$ 3.20 & 80.94 $\pm$ 9.77 & \bfseries 55.16 $\pm$ 0.97  \\
        \hline WRGAT  &83.62 $\pm$ 5.50&86.98 $\pm$ 3.78&36.53 $\pm$ 0.77&48.85 $\pm$ 0.78&65.24 $\pm$ 0.87&81.62 $\pm$ 3.90 & 54.76 ± 1.12  \\
        +CSA-I &83.69 $\pm$ 5.63&87.23 $\pm$ 3.94&36.55 $\pm$ 0.93&\bfseries 49.46 $\pm$ 0.74&65.36 $\pm$ 1.05&81.88 $\pm$ 3.93 & 54.86 ± 1.31 \\
        +CSA-II  &83.76 $\pm$ 5.61&87.02 $\pm$ 3.55&36.47 $\pm$ 0.74&48.93 $\pm$ 0.89&65.33 $\pm$ 0.92&\bfseries 82.76 $\pm$ 3.67 & 54.97 ± 1.23\\
        +CSA-III &\bfseries 84.88 $\pm$ 5.23&\bfseries 87.86 $\pm$ 3.77&\bfseries 36.89 $\pm$ 0.72&49.43 $\pm$ 0.88&\bfseries 66.02 $\pm$ 1.01&82.43 $\pm$ 4.00 & \bfseries 55.33 ± 1.18  \\
			
\bottomrule
		\end{tabular}
	\end{adjustbox}
 	\caption{Classification accuracy on heterophily datasets. CSA-I, CSA-II, and CSA-III indicate our three counterfactual schemes respectively. 
	}\label{tab:res}
\end{table*}

Consider a simple case where we conduct the node classification task with a standard $L$-layer GAT. For each layer $l$, we have the node representations $X^{l-1} \in \mathbb{R}^{n \times d^{l-1}}$ from the previous layer as input. Then, we perform feature aggregation and updating with factual attention map $A^l$ to obtain the factual output feature $X^{l}=f(X^{l-1},A^{l})$. Similarly, when we intervene the attention maps of layer $l$ (e.g., assigning \textit{dummy} values using $do(\cdot)$ operation), we can get a counterfactual output feature $\hat{X}^{l}$. We further employ a learnable matrix $\mathcal{W}^l \in \mathbb{R}^{c \times d^{l}}$ ($c$ denotes the number of classes) to get the node's factual predicted label $Y_{pred}^{l}$ and counterfactual predicted label $\hat{Y}_{pred}^{l}$ using the corresponding features from layer $l$. Therefore, the causal effect of attention at layer $l$ is obtained as: $Y_{pred}^{l} - \hat{Y}_{pred}^{l}$.
To this end, we can use the causal effect as a supervision signal to explicitly guide the attention learning process. The new objective of the CSA-assisted GAT model can be formulated as:
\begin{equation}\label{equ:training}
	\mathcal{L} \!\! = \!\!  \sum_{l} \lambda_l \mathcal{L}_{ce}(Y_\text{effect}^{l}, y) \!\!+\!\!  \mathcal{L}_\text{others},
\end{equation}
where $y$ is the ground-truth label,  $\mathcal{L}_{ce}$ is the cross-entropy loss, $\lambda_l$ is the coefficient to balance training, and $ \mathcal{L}_\text{others}$ represents the original objective such as standard classification loss.
Note that Equation.\eqref{equ:training} is a general form of CSA where we compute additional losses for each GAT layer to supervise attention directly. However in practice it is not necessary, and we found that simply selecting one or two layers is enough for CSA to bring satisfying performance improvement.

Moreover, since our aim is to boost the quality of attention, it is not necessary to estimate the correct causal effect of attention using \emph{dummy} values. Instead, a strong counterfactual baseline might 
even be helpful for the attention quality improvement. We hereby further propose three heuristic counterfactual schemes and test them in our experiments. We note that the exact form of how counterfactual is achieved is not limited, and our goal here is just to set the ball rolling.

\noindent\textbf{Scheme I:} In the first scheme, we utilize the uniform distribution to generate the counterfactual attention map. Specifically, the counterfactual attention is produced by 
\begin{align}
     \hat{a} \sim U(e,f),
\end{align}
where $e$ and $f$ are the lower and upper boundaries. In this case, the generated counterfactual could vary from very bad (i.e., mapping all unrelated neighbors) to very good (i.e., mapping all meaningful neighbors). 
This is a similar process to the \textit{Randomized Controlled Trial} \cite{stolberg2004randomized} where all possible treatments are enumerated. We hope that maximizing causal effects computed over all possible treatments can lead to a robust improvement of attention. 

\noindent\textbf{Scheme II:} Scheme I is easy and straightforward to apply. However due to its randomness, a possible concern is that if most of the generated counterfactual attentions are inferior to the factual one, then we will only have very small gradient on attention improvements. Therefore we are actually motivated to find  ``better'' counterfactuals to spur the factual one to evolve. Heuristically, given that MLP is a strong baseline on several datasets (e.g., Texas, Cornell, and Wisconsin), we employ an identical mapping to  
generate the counterfactual attention, which only attends to the ego-node instead of neighbors. Specifically, the counterfactual attention map is equal to the identity matrix $I$:
\begin{align}
     \hat{a}\sim I
\end{align}

\noindent\textbf{Scheme III:} Our last schema can be considered as an extension of Schema II. Since the fast development of GAT family has introduced to us some variants that already outperform MLP in many datasets, using counterfactuals derived from the behavior of MLP does not seem to be a wise choice for these GAT variants to improve their attentions. Inspired by the self-boosting concept \cite{pi2016self} widely used in machine learning, we leverage the historical attention map as the counterfactual to urge the factual one keep refining itself. The specific formulation is written as follows:
\begin{align}
     \hat{a}\sim A_{hist},
\end{align}
where $A_{hist}$ denotes the historical attention map (e.g., the attention map from the last update iteration).




 \section{Experiment}
 In this section, we conduct extensive node classification experiments to evaluate the performance of CSA. Specifically, we 1) validate the effectiveness of CSA on three popular GAT variants using a wide range of datasets, including both homophily and heterophily scenarios; 2) compare CSA with other attention improvement baselines of GAT to show the superiority of our method; 3) show that CSA can induce better attention that improve the robustness of GAT; 4) test CSA's sensitivity to hyper-parameters; 5) analyze the influences of CSA in feature space; and 6) examine the performances of some special cases of CSA.

\subsection{Datasets}

For heterophily scenario, we select seven standard benchmark datasets: Wisconsin, Cornell, Texas, Actor, Squirrel, Chameleon, and Crocodile.  Wisconsin, Cornell, and Texas collected by Carnegie Mellon University are published in WebKB\footnote{http://www.cs.cmu.edu/~webkb/}. Actor \cite{pei2020geom} is the actor-only  subgraph sampling from a film-director-actor-writer network. Squirrel, Chameleon, and Crocodile are datasets
collected from the English Wikipedia \cite{rozemberczki2021multi}. We summarize the basic attributions for each dataset in \tableref{tab:res}. $\mathcal{H}(\mathcal{G})$ is the  homophily ratio \cite{zhu2020beyond}, where $\mathcal{H}(\mathcal{G})\rightarrow 1$ represents extreme homophily and vice versa.

For homophily scenario, two large datasets released by Open Graph Benchmark (OGB)\footnote{https://ogb.stanford.edu/} \cite{hu2020open}: ogbn-products and ogbn-arxiv, are included in our experiments, together with two  small-scale homophily graph datasets: Cora and Citeseer \cite{mccallum2000automating}. Similarly, the attribution of the dataset is summarized in \tableref{tab:res2}.

\subsection{Experimental Setup}
We employ popular node classification models in our experiments as the baselines:  GCN \cite{kipf2016semi}, GAT \cite{velivckovic2017graph}, SGCN \cite{wu2019simplifying}, FAGCN \cite{bo2021beyond}, GPR-GNN \cite{chien2020adaptive}, H2GCN \cite{zhu2020beyond}, WRGAT \cite{suresh2021breaking}, APPNP \cite{gasteiger2018predict} and UniMP \cite{shi2020masked}. 
We also present the performance of MLPs, serving as a strong non-graph-based baseline. Due to page limit, we only select four models: GAT, FAGCN, WRGAT and UniMP to examine the effectiveness of CSA. These models ranges from the classic ones to the latest ones, and are considered as representatives for state-of-the-art node classification models. One thing to be noted here is that for all these models, we implement CSA only in their first layers to avoid excessive computational cost.



In our experiments, each GNN is run with the best hyperparameters if provided.
We set the same random seed for each model and dataset for reproducibility. The reported results are in the form of mean and standard deviation, calculated from 10 random node splits (the ratio of train/validation/test is 48\%/32\%/20\% from \cite{pei2020geom}). 
Our experiments are conducted on a GPU server with eight NVIDIA DGX A100 graphics cards, and the codes are implemented using Cuda Toolkit 11.5, PyTorch 1.8.1 and torch-geometric 2.0.1.

\subsection{Performance Analysis}

Table \ref{tab:res} and Table \ref{tab:res2} provide the test accuracy of different GNNs in different variants of CSA in the
supervised node classification task.  A graph’s homophily level is the average of its
nodes’ homophily levels. CSA achieves the best in terms of the vanilla one across all datasets.  In particular, the highest improved datasets are Texas, Wisconsin, and Cornell. By observing the performance of MLPs, we can see that the common ground of those three datasets contains distinguishable features and a large part of non-homophilous edges. In the meanwhile, the performance of  CSA is 
proportional to the modeling ability.  The mechanism behind CSA is to extend the causal effect between nodes representation and final prediction. Therefore,  CSA owns limited performance when the node's representations are chaotic. Our experiments highlight that I) The model, which is already better than MLP, does not improve much in CSA-II, while CSA-III improves it relatively more. This is because in those datasets, the graph structure can provide meaningful information, so that CSA-III have more advantages.  II) The dataset, which has distinctive features indicated by the performance of MLPs, is more satisfied CSA-II. Similarly, in this scene, the features can be more informative. III) The random strategy (CSA-I) relatively inferior to others, since the distribution is hard to control and tend to generate worst attention map, whereby the regularization is vanished.


\begin{table}[]
\centering

\begin{adjustbox}{width=0.47\textwidth}
\begin{tabular}{l|cccc}
\toprule
\  Models  & \bfseries Cora & \bfseries CiteSeer &\bfseries ogbn-products & \bfseries ogbn-arxiv \\

$\mathcal{H}(\mathcal{G})$ &0.81  &0.74 &0.81      &0.66 	 \\
\textbf{\#Nodes} &			2,708&3,327&2,449,029 &169,343	\\
\textbf{\#Edges} &  		5,278&4,467&61,859,140&1,166,243	\\
\textbf{\#Classes} & 		7    &7    &47        &40	\\
\textbf{\#Features} &		1433 &3703 &100       &128	 \\

\hline GAT & 86.21 $\pm$ 0.78 & 75.73 $\pm$ 1.23 &77.02 $\pm$ 0.63   & 70.96  $\pm$  0.14 \\
+CSA-I  & 86.16 $\pm$ 0.95 & 76.81 $\pm$ 1.29 &77.28 $\pm$ 0.69   & 71.08  $\pm$ 0.14\\
+CSA-II  & 86.89 $\pm$ 0.64 & 76.53 $\pm$ 1.18 &77.44 $\pm$ 0.63   & 71.05  $\pm$ 0.14\\
+CSA-III  & \bfseries 87.86 $\pm$ 0.87 & \bfseries 77.72 $\pm$ 1.25 & \bfseries 78.36 $\pm$ 0.72   & \bfseries 71.20  $\pm$ 0.16\\
\hline UniMP & 86.89 ± 0.90& 75.14 ± 0.68 & 81.37 ± 0.47 & 72.92 ± 0.10 \\
+CSA-I & 87.47 ± 0.87& 75.89 ± 0.73 & 81.55 ± 0.62 & 72.94 ± 0.10 \\
+CSA-II & 85.62 ± 0.73 & 75.87 ± 0.72 & 81.39 ± 0.47 & 72.96 ± 0.10 \\
+CSA-III & \bfseries 88.64 ± 1.28& \bfseries 77.61 ± 0.82 & \bfseries 82.24 ± 0.63 & \bfseries73.08 ± 0.11 \\
\bottomrule
\end{tabular}
\end{adjustbox}
\caption{Classification accuracy on homophily datasets.}\label{tab:res2}
\end{table}

\begin{table}[t]
\centering

\begin{adjustbox}{width=0.45\textwidth}
\begin{tabular}{l|ccc}
\toprule
\  Models  & \bfseries Texas& \bfseries Cornell& 
\bfseries ogbn-arxiv 
\\
\hline GAT & $55.21 \pm 5.70$ & $61.89 \pm 6.08$ & $70.96 \pm 0.14$\\
+CSA (Last) & $57.83 \pm 4.65$ & $63.52 \pm 5.34$ & $71.08 \pm 0.15$\\
+CSA (Pure) & $55.79 \pm 5.05$ & $61.97 \pm 5.18$ & $70.96 \pm 0.14$\\
+CSA (Ours) & \bfseries 58.21 $\pm$ 4.79 & \bfseries 64.26 $\pm$ 5.21 & \bfseries 71.20 $\pm$ 0.16\\
 \bottomrule
\end{tabular}
\end{adjustbox}

\caption{Comparison with the heuristic causal strategies.}\label{tab:res3}
\end{table}

\begin{figure}[h]
	\centering
    \subfigure[Texas]{
		\label{fig:difficult_pattern}
		\includegraphics[width=0.45\linewidth]{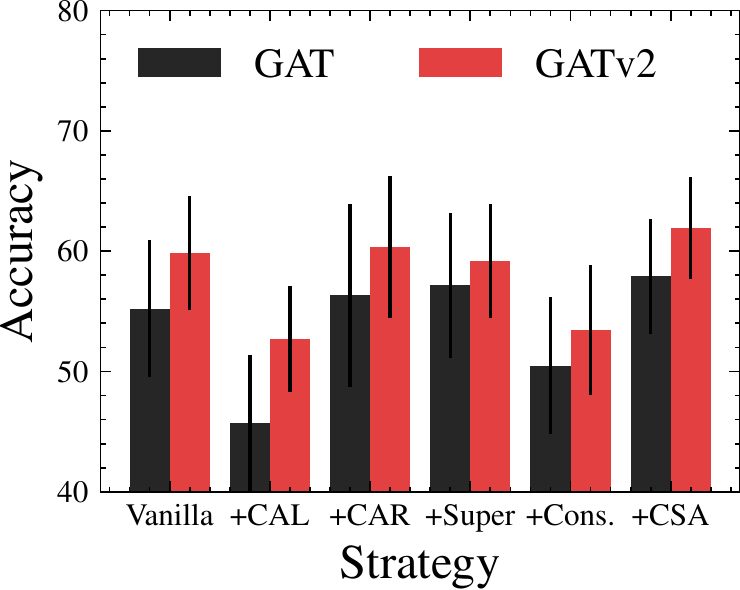}
	}
    \subfigure[Cornell]{
		\label{fig:difficult_pattern}
		\includegraphics[width=0.45\linewidth]{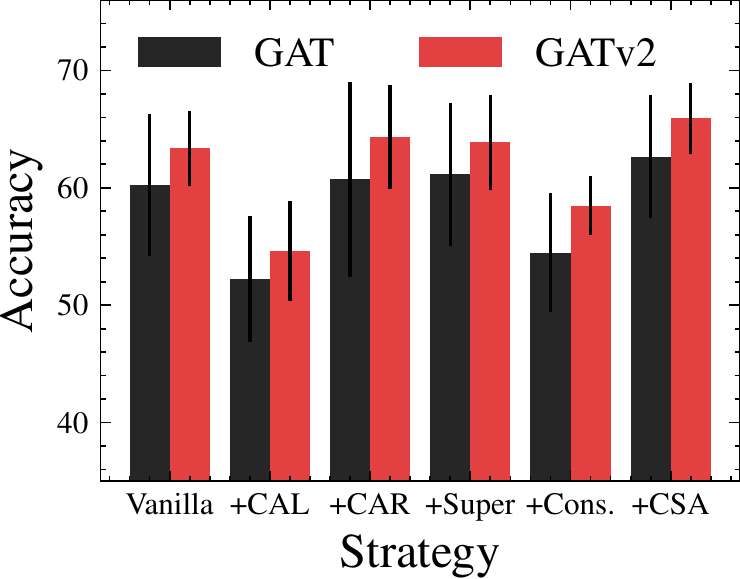}
	}

     \subfigure[Cora]{
		\label{fig:difficult_pattern}
		\includegraphics[width=0.45\linewidth]{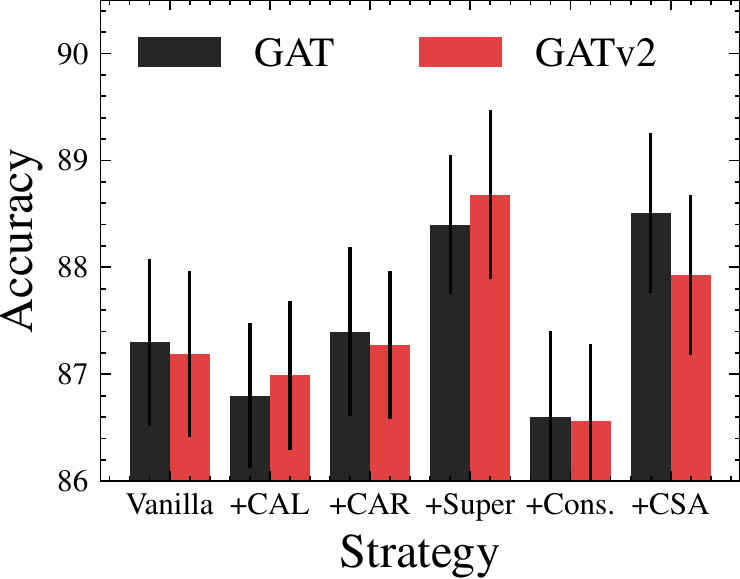}
	}
    \subfigure[CiteSeer]{
		\label{fig:difficult_pattern}
		\includegraphics[width=0.45\linewidth]{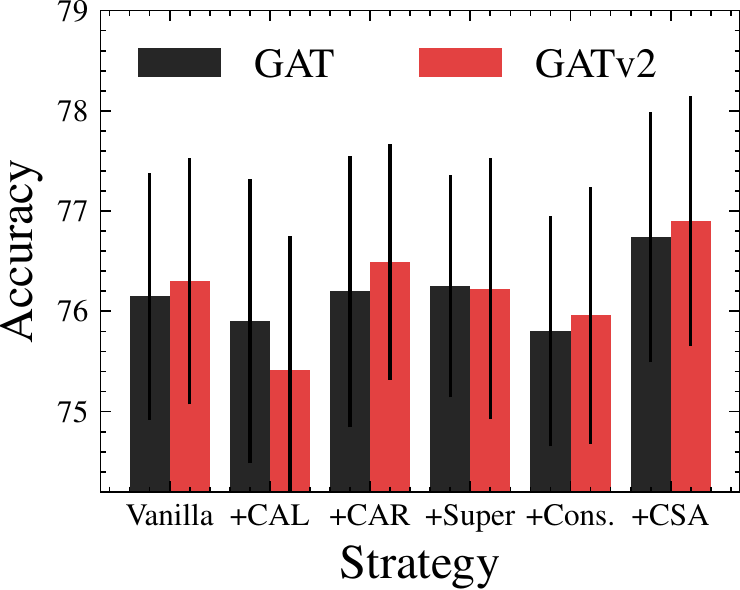}
	}
	\caption{Comparison with different GATs promotion strategies. }
	\label{fig:compare}
\end{figure}
\vspace{-0.3cm}


\subsection{Comparison with Attention Promotion Baselines }
We here apply multiple attention promotion baselines:  CAL \cite{sui2022causal}, CAR \cite{wu2022causally}, Super \cite{kim2022find}, Constraint interventions \cite{wang2019improving} and the result is shown in \figref{fig:compare}. Among them, CAL is a method for graph property prediction that
relies on an alternative formulation of causal attention with interventions on implicit representations.
We adapted CAL for node classification by removing its final pooling layer. Super is well-known as SuperGAT, a method seeking to constrain node feature vectors through a semi-supervised edge prediction task.  CAR aligns the attention mechanism with the causal effects of active interventions on graph connectivity in a scalable manner.   Constraint method has two auxiliary losses: graph structure-based constraint  and class boundary constraint. The results on four datasets are shown in \figref{fig:compare}. While CAL, CAR, and CSA have related goals of enhancing graph attention using concepts
from causal theory, CAL uses abstract perturbations on graph representation to perform causal interventions, and  CAR employs an edge intervention strategy that enables causal effects to be computed
scalable, while our method does not exert any assumptions and constraints on GATs, compared with CAL and CAR. Therefore, CSA tends to own good generalization ability. 
In terms of SuperGAT and Constraint method, since there is a trade-off between node classification and regularization. For example,  SuperGAT implies that it is hard to learn the relational importance of edges by simply optimizing graph attention for link prediction.  

\begin{figure}[t]
	\centering
        \includegraphics[width=1\linewidth]{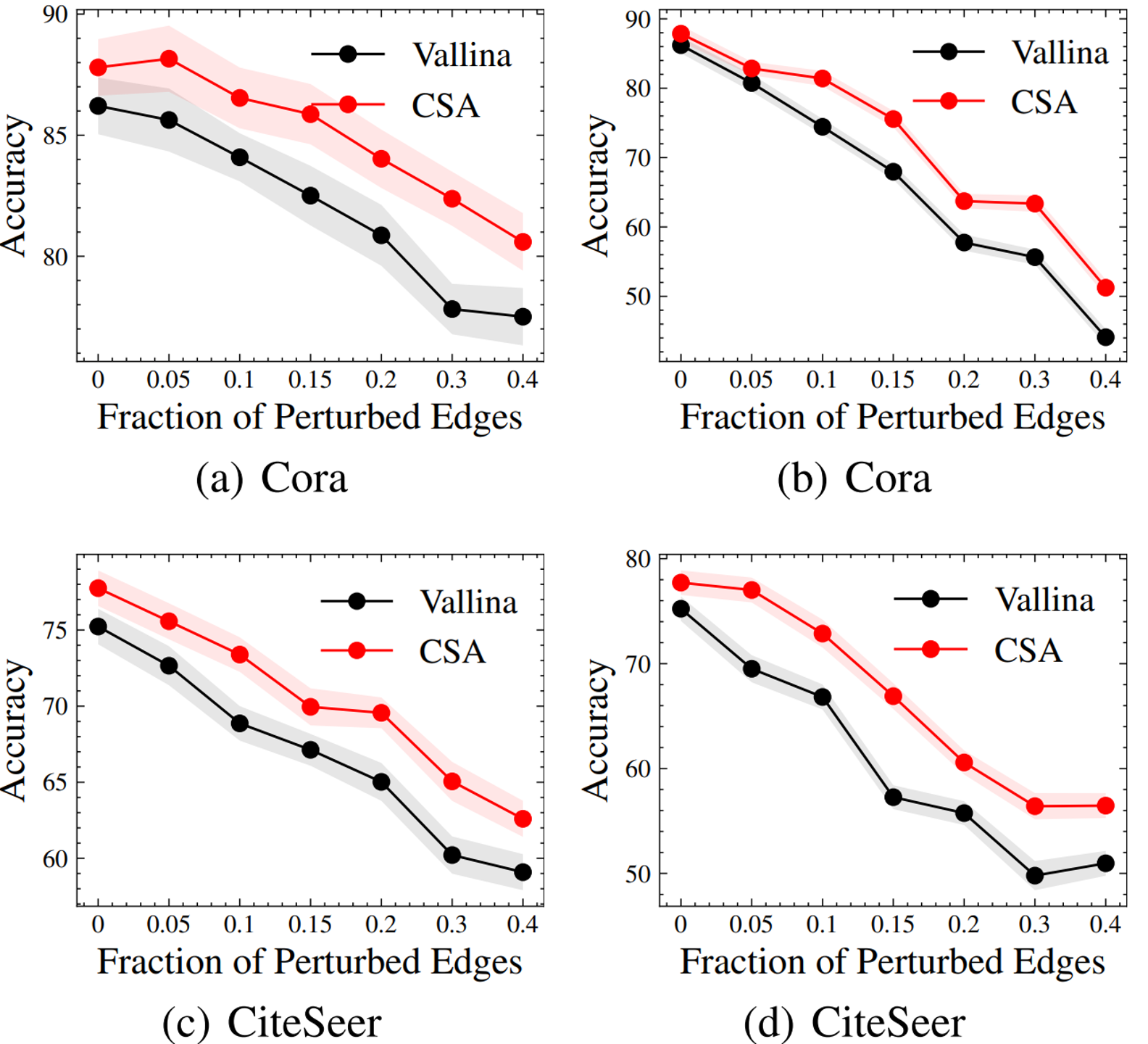}
	\caption{The robust performance on the node and edge perturbations. }
	\label{fig:robust}
\end{figure}

\begin{figure}[h]
	\centering
    \subfigure[Texas]{
		\label{fig:difficult_pattern}
		\includegraphics[width=0.45\linewidth]{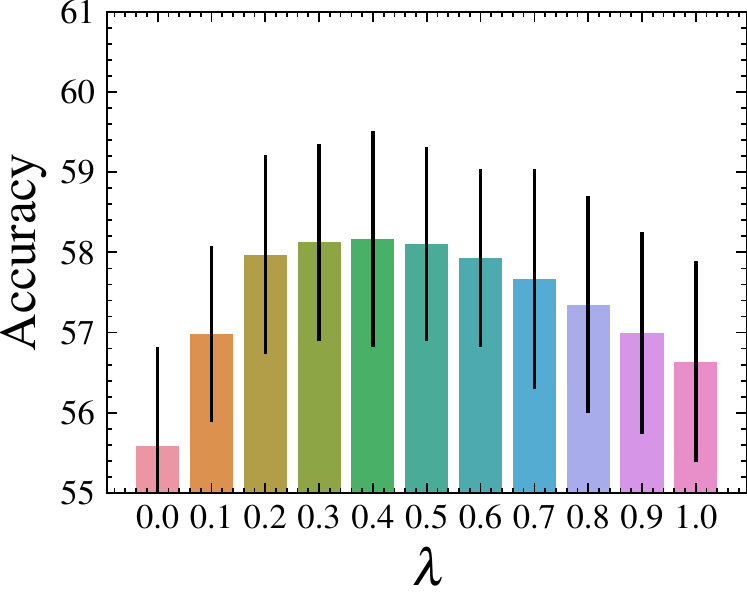}
	}
    \subfigure[Cora]{
		\label{fig:difficult_pattern}
		\includegraphics[width=0.45\linewidth]{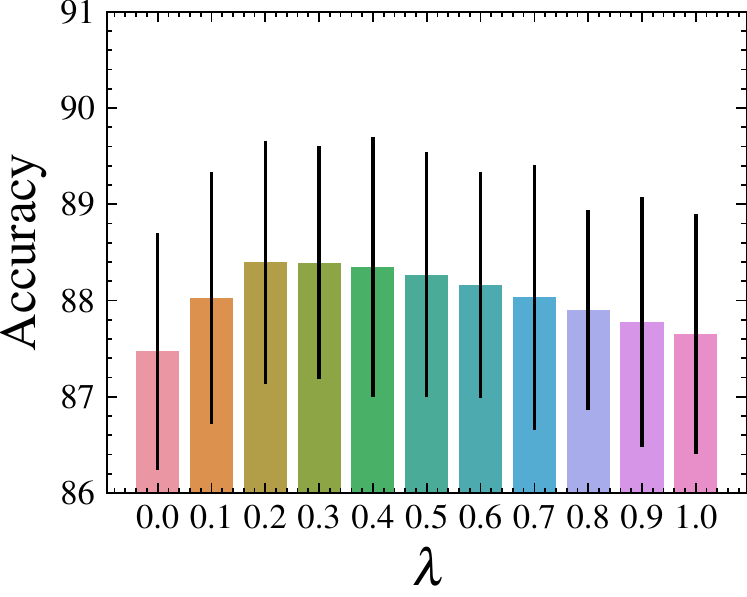}
	}
	\caption{Hyper-parameter analysis on GAT. }
	\label{fig:lambda}
\end{figure}




\subsection{CSA Provides Robustness}
In this section, we systematically study the performance of CSA on  robustness against the input perturbations including feature and edge perturbations. Following \cite{stadler2021graph},  we conduct node-level feature perturbations by replacing them with the noise sampled from the Bernoulli distribution with $p=0.5$ and edge perturbations by 
stochastically generating the edges. According to the performance shown in \figref{fig:robust}, CSA produces robust performance on input perturbations. \figref{fig:robust} demonstrate that CSA in higher noise situations achieves more robust results
than in lower noise scenes on both node and edge perturbations with perturbation percentages
ranging from $0\%$ to $40\%$.

\subsection{Hyper-parameter Analysis}
We analyze the sensitivity of $\lambda$ and plot node classification performance in \figref{fig:lambda}. For $\lambda$, there is a specific range that maximizes test performance in all datasets. The performance in Texas is the highest when $\lambda$ is $0.4$, but the difference is relatively small compared to Cora. We observe that there is an optimal level of causal supervision for each dataset, and using too large $\lambda$ degrades node classification performance. Since Cora owns friendly neighbors, its performance is less sensitive than Texas. Based on this, we can also see that Texas relatively needs a larger regularization.   

\subsection{Pairwise Distance among Classes}
\begin{figure}[t]
	\centering
    \subfigure[Texas]{
	\label{}
    \includegraphics[width=0.45\linewidth]{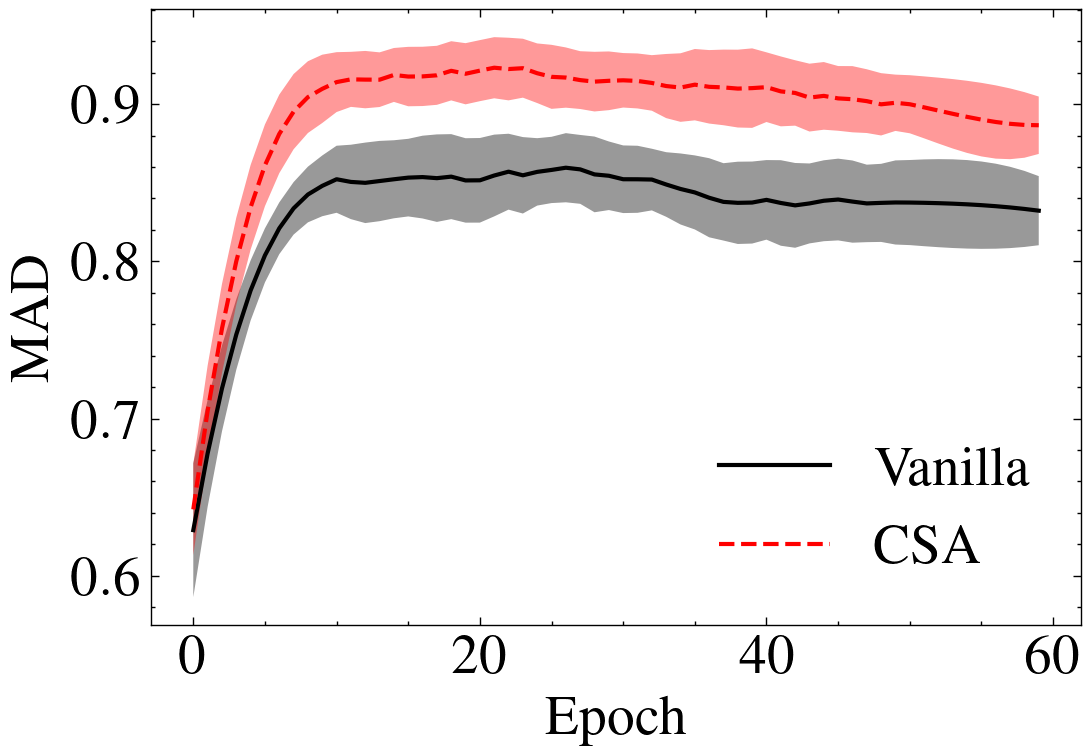}
	}
    \subfigure[Cornell]{
		\label{}
		\includegraphics[width=0.45\linewidth]{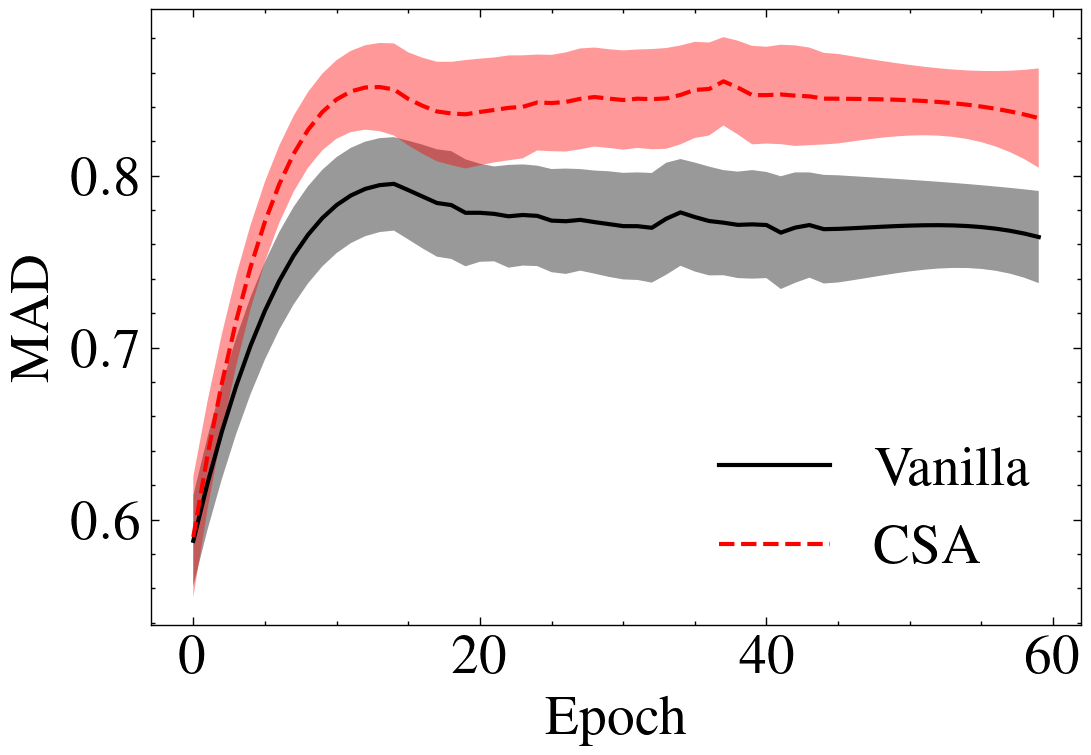}
	}
	\caption{ Mean Average Distance among node representations of Last GAT layer.}
 \vspace{-0.3cm}
	\label{fig:distance}
\end{figure}

 To further evaluate whether the good performance of CSA can be contributed to the mitigation of lacking supervision, we visualize the pairwise distance of the node representations among classes learned by CSA and vanilla GAT. Following \cite{stadler2021graph}, we calculate the  Mean Average Distance (MAD) with cosine distance among node representations for the last layer.  The larger the MAD is, the better the node representations are. Results are reported in \figref{fig:distance}. It can be observed that the node representations learned by CSA keep having a large distance throughout the optimization process, indicating relief of lacking supervision issues. On the contrary, GAT suffers from severely indistinguishable representations of nodes.

\subsection{Comparison with Heuristic Causal Strategies.}
To validate the effectiveness of CSA, we compare it with two heuristic causal designs (Last and Pure) that 1) directly estimate the total causal effect by subtracting between the model's and the counterfactual's output in the final layer; 2) replace the attention with the static aggregate weights (i.e., each node allocates the same weight).  The results are shown in Table \ref{tab:res3}. We observe that their performance 
outperforms vanilla one, but is still inferior to ours. In terms of Last, the major difference is whether to explicitly estimate causal effect or not. In our framework, we plug the MLPs into the hidden layers to precisely estimate the causal effect for each layer. Regarding Pure, our strategy can provide more strong baselines, leading to better regularization.

\section{Conclusion}
We introduced CSA, a counterfactual-based regularization scheme that can be applied
to graph attention architectures. Unlike other causal approaches, we first built the causal graph of GATs in a general way and do not impose any assumptions and constraints on GATs. Subsequently, we introduce an efficient scheme to directly estimate the causal effect of attention in hidden layers. Applying
it to both homophilic and heterophilic node-classification tasks, we found accuracy improvements
and robustness in almost all circumstances. We performed three variants of counterfactual attention strategies and found that they can adapt to different situations, respectively.

\section*{Contribution Statement}
Hongjun Wang and Jiyuan Chen contributed equally to the work. Lun Du and Xuan Song are the corresponding authors.
This work is done during Hongjun Wang’s internship at Microsoft Research Asia.

{\small

	\bibliographystyle{named}
	\bibliography{citation}

\begin{thebibliography}{}

\bibitem[\protect\citeauthoryear{Battaglia \bgroup \em et al.\egroup
  }{2016}]{battaglia2016interaction}
Peter Battaglia, Razvan Pascanu, Matthew Lai, Danilo Jimenez~Rezende, et~al.
\newblock Interaction networks for learning about objects, relations and
  physics.
\newblock {\em Advances in neural information processing systems}, 29, 2016.

\bibitem[\protect\citeauthoryear{Bo \bgroup \em et al.\egroup
  }{2021}]{bo2021beyond}
Deyu Bo, Xiao Wang, Chuan Shi, and Huawei Shen.
\newblock Beyond low-frequency information in graph convolutional networks.
\newblock In {\em Proceedings of the AAAI Conference on Artificial
  Intelligence}, volume~35, pages 3950--3957, 2021.

\bibitem[\protect\citeauthoryear{Brody \bgroup \em et al.\egroup
  }{2021}]{brody2021attentive}
Shaked Brody, Uri Alon, and Eran Yahav.
\newblock How attentive are graph attention networks?
\newblock {\em arXiv preprint arXiv:2105.14491}, 2021.

\bibitem[\protect\citeauthoryear{Chien \bgroup \em et al.\egroup
  }{2020}]{chien2020adaptive}
Eli Chien, Jianhao Peng, Pan Li, and Olgica Milenkovic.
\newblock Adaptive universal generalized pagerank graph neural network.
\newblock {\em arXiv preprint arXiv:2006.07988}, 2020.

\bibitem[\protect\citeauthoryear{Cinelli \bgroup \em et al.\egroup
  }{2019}]{cinelli2019sensitivity}
Carlos Cinelli, Daniel Kumor, Bryant Chen, Judea Pearl, and Elias Bareinboim.
\newblock Sensitivity analysis of linear structural causal models.
\newblock In {\em International conference on machine learning}, pages
  1252--1261. PMLR, 2019.

\bibitem[\protect\citeauthoryear{Defferrard \bgroup \em et al.\egroup
  }{2016}]{defferrard2016convolutional}
Micha{\"e}l Defferrard, Xavier Bresson, and Pierre Vandergheynst.
\newblock Convolutional neural networks on graphs with fast localized spectral
  filtering.
\newblock {\em Advances in neural information processing systems}, 29, 2016.

\bibitem[\protect\citeauthoryear{Du \bgroup \em et al.\egroup
  }{2022}]{du2022gbk}
Lun Du, Xiaozhou Shi, Qiang Fu, Xiaojun Ma, Hengyu Liu, Shi Han, and Dongmei
  Zhang.
\newblock Gbk-gnn: Gated bi-kernel graph neural networks for modeling both
  homophily and heterophily.
\newblock In {\em Proceedings of the ACM Web Conference 2022}, pages
  1550--1558, 2022.

\bibitem[\protect\citeauthoryear{Feng \bgroup \em et al.\egroup
  }{2021}]{feng2021should}
Fuli Feng, Weiran Huang, Xiangnan He, Xin Xin, Qifan Wang, and Tat-Seng Chua.
\newblock Should graph convolution trust neighbors? a simple causal inference
  method.
\newblock In {\em Proceedings of the 44th International ACM SIGIR Conference on
  Research and Development in Information Retrieval}, pages 1208--1218, 2021.

\bibitem[\protect\citeauthoryear{Gao and Ji}{2019}]{gao2019graph}
Hongyang Gao and Shuiwang Ji.
\newblock Graph representation learning via hard and channel-wise attention
  networks.
\newblock In {\em Proceedings of the 25th ACM SIGKDD International Conference
  on Knowledge Discovery \& Data Mining}, pages 741--749, 2019.

\bibitem[\protect\citeauthoryear{Gasteiger \bgroup \em et al.\egroup
  }{2018}]{gasteiger2018predict}
Johannes Gasteiger, Aleksandar Bojchevski, and Stephan G{\"u}nnemann.
\newblock Predict then propagate: Graph neural networks meet personalized
  pagerank.
\newblock {\em arXiv preprint arXiv:1810.05997}, 2018.

\bibitem[\protect\citeauthoryear{Gilmer \bgroup \em et al.\egroup
  }{2017}]{gilmer2017neural}
Justin Gilmer, Samuel~S Schoenholz, Patrick~F Riley, Oriol Vinyals, and
  George~E Dahl.
\newblock Neural message passing for quantum chemistry.
\newblock In {\em International conference on machine learning}, pages
  1263--1272. PMLR, 2017.

\bibitem[\protect\citeauthoryear{Hoshen}{2017}]{hoshen2017vain}
Yedid Hoshen.
\newblock Vain: Attentional multi-agent predictive modeling.
\newblock {\em Advances in Neural Information Processing Systems}, 30, 2017.

\bibitem[\protect\citeauthoryear{Hou \bgroup \em et al.\egroup
  }{2022}]{hou2022measuring}
Yifan Hou, Jian Zhang, James Cheng, Kaili Ma, Richard~TB Ma, Hongzhi Chen, and
  Ming-Chang Yang.
\newblock Measuring and improving the use of graph information in graph neural
  networks.
\newblock {\em arXiv preprint arXiv:2206.13170}, 2022.

\bibitem[\protect\citeauthoryear{Hu \bgroup \em et al.\egroup
  }{2020}]{hu2020open}
Weihua Hu, Matthias Fey, Marinka Zitnik, Yuxiao Dong, Hongyu Ren, Bowen Liu,
  Michele Catasta, and Jure Leskovec.
\newblock Open graph benchmark: Datasets for machine learning on graphs.
\newblock {\em Advances in neural information processing systems},
  33:22118--22133, 2020.

\bibitem[\protect\citeauthoryear{Huang \bgroup \em et al.\egroup
  }{2023}]{huang2023robust}
Jincheng Huang, Lun Du, Xu~Chen, Qiang Fu, Shi Han, and Dongmei Zhang.
\newblock Robust mid-pass filtering graph convolutional networks.
\newblock In {\em Proceedings of the ACM Web Conference 2023}, pages 328--338,
  2023.

\bibitem[\protect\citeauthoryear{Kim and Oh}{2020}]{kim2020find}
Dongkwan Kim and Alice Oh.
\newblock How to find your friendly neighborhood: Graph attention design with
  self-supervision.
\newblock In {\em ICLR}, 2020.

\bibitem[\protect\citeauthoryear{Kim and Oh}{2022}]{kim2022find}
Dongkwan Kim and Alice Oh.
\newblock How to find your friendly neighborhood: Graph attention design with
  self-supervision.
\newblock {\em arXiv preprint arXiv:2204.04879}, 2022.

\bibitem[\protect\citeauthoryear{Kipf and Welling}{2016}]{kipf2016semi}
Thomas~N Kipf and Max Welling.
\newblock Semi-supervised classification with graph convolutional networks.
\newblock {\em arXiv preprint arXiv:1609.02907}, 2016.

\bibitem[\protect\citeauthoryear{Knyazev \bgroup \em et al.\egroup
  }{2019}]{knyazev2019understanding}
Boris Knyazev, Graham~W Taylor, and Mohamed Amer.
\newblock Understanding attention and generalization in graph neural networks.
\newblock {\em Advances in neural information processing systems}, 32, 2019.

\bibitem[\protect\citeauthoryear{Kumar \bgroup \em et al.\egroup
  }{2010}]{kumar2010self}
M~Pawan Kumar, Benjamin Packer, and Daphne Koller.
\newblock Self-paced learning for latent variable models.
\newblock In {\em NIPS}, volume~1, page~2, 2010.

\bibitem[\protect\citeauthoryear{Lee \bgroup \em et al.\egroup
  }{2019}]{lee2018attention}
John~Boaz Lee, Ryan~A Rossi, Sungchul Kim, Nesreen~K Ahmed, and Eunyee Koh.
\newblock Attention models in graphs: A survey.
\newblock {\em ACM Transactions on Knowledge Discovery from Data (TKDD)},
  13(6):1--25, 2019.

\bibitem[\protect\citeauthoryear{Li \bgroup \em et al.\egroup
  }{2022}]{li2022finding}
Xiang Li, Renyu Zhu, Yao Cheng, Caihua Shan, Siqiang Luo, Dongsheng Li, and
  Weining Qian.
\newblock Finding global homophily in graph neural networks when meeting
  heterophily.
\newblock {\em arXiv preprint arXiv:2205.07308}, 2022.

\bibitem[\protect\citeauthoryear{Liu \bgroup \em et al.\egroup
  }{2021}]{liu2021non}
Meng Liu, Zhengyang Wang, and Shuiwang Ji.
\newblock Non-local graph neural networks.
\newblock {\em IEEE Transactions on Pattern Analysis and Machine Intelligence},
  2021.

\bibitem[\protect\citeauthoryear{McCallum \bgroup \em et al.\egroup
  }{2000}]{mccallum2000automating}
Andrew~Kachites McCallum, Kamal Nigam, Jason Rennie, and Kristie Seymore.
\newblock Automating the construction of internet portals with machine
  learning.
\newblock {\em Information Retrieval}, 3:127--163, 2000.

\bibitem[\protect\citeauthoryear{Pearl}{2009}]{pearl2009causality}
Judea Pearl.
\newblock {\em Causality}.
\newblock Cambridge university press, 2009.

\bibitem[\protect\citeauthoryear{Pearl}{2014}]{pearl2014interpretation}
Judea Pearl.
\newblock Interpretation and identification of causal mediation.
\newblock {\em Psychological methods}, 19(4):459, 2014.

\bibitem[\protect\citeauthoryear{Pearl}{2022}]{pearl2022direct}
Judea Pearl.
\newblock Direct and indirect effects.
\newblock In {\em Probabilistic and Causal Inference: The Works of Judea
  Pearl}, pages 373--392. 2022.

\bibitem[\protect\citeauthoryear{Pei \bgroup \em et al.\egroup
  }{2020}]{pei2020geom}
Hongbin Pei, Bingzhe Wei, Kevin Chen-Chuan Chang, Yu~Lei, and Bo~Yang.
\newblock Geom-gcn: Geometric graph convolutional networks.
\newblock {\em arXiv preprint arXiv:2002.05287}, 2020.

\bibitem[\protect\citeauthoryear{Pi \bgroup \em et al.\egroup
  }{2016}]{pi2016self}
Te~Pi, Xi~Li, Zhongfei Zhang, Deyu Meng, Fei Wu, Jun Xiao, Yueting Zhuang,
  et~al.
\newblock Self-paced boost learning for classification.
\newblock In {\em IJCAI}, pages 1932--1938, 2016.

\bibitem[\protect\citeauthoryear{Ribeiro \bgroup \em et al.\egroup
  }{2016}]{ribeiro2016should}
Marco~Tulio Ribeiro, Sameer Singh, and Carlos Guestrin.
\newblock " why should i trust you?" explaining the predictions of any
  classifier.
\newblock In {\em Proceedings of the 22nd ACM SIGKDD international conference
  on knowledge discovery and data mining}, pages 1135--1144, 2016.

\bibitem[\protect\citeauthoryear{Rozemberczki \bgroup \em et al.\egroup
  }{2021}]{rozemberczki2021multi}
Benedek Rozemberczki, Carl Allen, and Rik Sarkar.
\newblock Multi-scale attributed node embedding.
\newblock {\em Journal of Complex Networks}, 9(2):cnab014, 2021.

\bibitem[\protect\citeauthoryear{Santoro \bgroup \em et al.\egroup
  }{2017}]{santoro2017simple}
Adam Santoro, David Raposo, David~G Barrett, Mateusz Malinowski, Razvan
  Pascanu, Peter Battaglia, and Timothy Lillicrap.
\newblock A simple neural network module for relational reasoning.
\newblock {\em Advances in neural information processing systems}, 30, 2017.

\bibitem[\protect\citeauthoryear{Shi \bgroup \em et al.\egroup
  }{2020}]{shi2020masked}
Yunsheng Shi, Zhengjie Huang, Shikun Feng, Hui Zhong, Wenjin Wang, and Yu~Sun.
\newblock Masked label prediction: Unified message passing model for
  semi-supervised classification.
\newblock {\em arXiv preprint arXiv:2009.03509}, 2020.

\bibitem[\protect\citeauthoryear{Stadler \bgroup \em et al.\egroup
  }{2021}]{stadler2021graph}
Maximilian Stadler, Bertrand Charpentier, Simon Geisler, Daniel Z{\"u}gner, and
  Stephan G{\"u}nnemann.
\newblock Graph posterior network: Bayesian predictive uncertainty for node
  classification.
\newblock {\em Advances in Neural Information Processing Systems},
  34:18033--18048, 2021.

\bibitem[\protect\citeauthoryear{Stolberg \bgroup \em et al.\egroup
  }{2004}]{stolberg2004randomized}
Harald~O Stolberg, Geoffrey Norman, and Isabelle Trop.
\newblock Randomized controlled trials.
\newblock {\em AJR Am J Roentgenol}, 183(6):1539--44, 2004.

\bibitem[\protect\citeauthoryear{Sui \bgroup \em et al.\egroup
  }{2022}]{sui2022causal}
Yongduo Sui, Xiang Wang, Jiancan Wu, Min Lin, Xiangnan He, and Tat-Seng Chua.
\newblock Causal attention for interpretable and generalizable graph
  classification.
\newblock In {\em Proceedings of the 28th ACM SIGKDD Conference on Knowledge
  Discovery and Data Mining}, pages 1696--1705, 2022.

\bibitem[\protect\citeauthoryear{Sun \bgroup \em et al.\egroup
  }{2021}]{sun2021scalable}
Chuxiong Sun, Hongming Gu, and Jie Hu.
\newblock Scalable and adaptive graph neural networks with self-label-enhanced
  training.
\newblock {\em arXiv preprint arXiv:2104.09376}, 2021.

\bibitem[\protect\citeauthoryear{Suresh \bgroup \em et al.\egroup
  }{2021}]{suresh2021breaking}
Susheel Suresh, Vinith Budde, Jennifer Neville, Pan Li, and Jianzhu Ma.
\newblock Breaking the limit of graph neural networks by improving the
  assortativity of graphs with local mixing patterns.
\newblock {\em arXiv preprint arXiv:2106.06586}, 2021.

\bibitem[\protect\citeauthoryear{Thekumparampil \bgroup \em et al.\egroup
  }{2018}]{thekumparampil2018attention}
Kiran~K Thekumparampil, Chong Wang, Sewoong Oh, and Li-Jia Li.
\newblock Attention-based graph neural network for semi-supervised learning.
\newblock {\em arXiv preprint arXiv:1803.03735}, 2018.

\bibitem[\protect\citeauthoryear{VanderWeele}{2016}]{vanderweele2016explanation}
Tyler~J VanderWeele.
\newblock Explanation in causal inference: developments in mediation and
  interaction.
\newblock {\em International journal of epidemiology}, 45(6):1904--1908, 2016.

\bibitem[\protect\citeauthoryear{Vaswani \bgroup \em et al.\egroup
  }{2017}]{vaswani2017attention}
Ashish Vaswani, Noam Shazeer, Niki Parmar, Jakob Uszkoreit, Llion Jones,
  Aidan~N Gomez, {\L}ukasz Kaiser, and Illia Polosukhin.
\newblock Attention is all you need.
\newblock {\em Advances in neural information processing systems}, 30, 2017.

\bibitem[\protect\citeauthoryear{Veli{\v{c}}kovi{\'c} \bgroup \em et al.\egroup
  }{2017}]{velivckovic2017graph}
Petar Veli{\v{c}}kovi{\'c}, Guillem Cucurull, Arantxa Casanova, Adriana Romero,
  Pietro Lio, and Yoshua Bengio.
\newblock Graph attention networks.
\newblock {\em arXiv preprint arXiv:1710.10903}, 2017.

\bibitem[\protect\citeauthoryear{Wang \bgroup \em et al.\egroup
  }{2019}]{wang2019improving}
Guangtao Wang, Rex Ying, Jing Huang, and Jure Leskovec.
\newblock Improving graph attention networks with large margin-based
  constraints.
\newblock {\em arXiv preprint arXiv:1910.11945}, 2019.

\bibitem[\protect\citeauthoryear{Wang \bgroup \em et al.\egroup
  }{2021}]{wang2021evolving}
Yujing Wang, Yaming Yang, Jiangang Bai, Mingliang Zhang, Jing Bai, Jing Yu,
  Ce~Zhang, Gao Huang, and Yunhai Tong.
\newblock Evolving attention with residual convolutions.
\newblock In {\em International Conference on Machine Learning}, pages
  10971--10980. PMLR, 2021.

\bibitem[\protect\citeauthoryear{Wang \bgroup \em et al.\egroup
  }{2022a}]{wang2022st}
Hongjun Wang, Jiyuan Chen, Zipei Fan, Zhiwen Zhang, Zekun Cai, and Xuan Song.
\newblock St-expertnet: A deep expert framework for traffic prediction.
\newblock {\em IEEE Transactions on Knowledge and Data Engineering}, 2022.

\bibitem[\protect\citeauthoryear{Wang \bgroup \em et al.\egroup
  }{2022b}]{wang2022easy}
Hongjun Wang, Jiyuan Chen, Tong Pan, Zipei Fan, Boyuan Zhang, Renhe Jiang,
  Lingyu Zhang, Yi~Xie, Zhongyi Wang, and Xuan Song.
\newblock Easy begun is half done: Spatial-temporal graph modeling with
  st-curriculum dropout.
\newblock {\em arXiv preprint arXiv:2211.15182}, 2022.

\bibitem[\protect\citeauthoryear{Wu \bgroup \em et al.\egroup
  }{2019}]{wu2019simplifying}
Felix Wu, Amauri Souza, Tianyi Zhang, Christopher Fifty, Tao Yu, and Kilian
  Weinberger.
\newblock Simplifying graph convolutional networks.
\newblock In {\em International conference on machine learning}, pages
  6861--6871. PMLR, 2019.

\bibitem[\protect\citeauthoryear{Wu \bgroup \em et al.\egroup
  }{2022a}]{wu2022causally}
Alexander~P Wu, Thomas Markovich, Bonnie Berger, Nils Hammerla, and Rohit
  Singh.
\newblock Causally-guided regularization of graph attention improves
  generalizability.
\newblock {\em arXiv preprint arXiv:2210.10946}, 2022.

\bibitem[\protect\citeauthoryear{Wu \bgroup \em et al.\egroup
  }{2022b}]{wu2022graph}
Shiwen Wu, Fei Sun, Wentao Zhang, Xu~Xie, and Bin Cui.
\newblock Graph neural networks in recommender systems: a survey.
\newblock {\em ACM Computing Surveys}, 55(5):1--37, 2022.

\bibitem[\protect\citeauthoryear{Xu \bgroup \em et al.\egroup
  }{2018}]{xu2018powerful}
Keyulu Xu, Weihua Hu, Jure Leskovec, and Stefanie Jegelka.
\newblock How powerful are graph neural networks?
\newblock {\em arXiv preprint arXiv:1810.00826}, 2018.

\bibitem[\protect\citeauthoryear{Ye and Ji}{2021}]{ye2021sparse}
Yang Ye and Shihao Ji.
\newblock Sparse graph attention networks.
\newblock {\em IEEE Transactions on Knowledge and Data Engineering}, 2021.

\bibitem[\protect\citeauthoryear{Ying \bgroup \em et al.\egroup
  }{2021}]{ying2021transformers}
Chengxuan Ying, Tianle Cai, Shengjie Luo, Shuxin Zheng, Guolin Ke, Di~He,
  Yanming Shen, and Tie-Yan Liu.
\newblock Do transformers really perform badly for graph representation?
\newblock {\em Advances in Neural Information Processing Systems},
  34:28877--28888, 2021.

\bibitem[\protect\citeauthoryear{You \bgroup \em et al.\egroup
  }{2020}]{you2020does}
Yuning You, Tianlong Chen, Zhangyang Wang, and Yang Shen.
\newblock When does self-supervision help graph convolutional networks?
\newblock In {\em international conference on machine learning}, pages
  10871--10880. PMLR, 2020.

\bibitem[\protect\citeauthoryear{Zhang and Chen}{2018}]{zhang2018link}
Muhan Zhang and Yixin Chen.
\newblock Link prediction based on graph neural networks.
\newblock In {\em NeurIPS}, pages 5165--5175, 2018.

\bibitem[\protect\citeauthoryear{Zhang \bgroup \em et al.\egroup
  }{2018}]{zhang2018gaan}
Jiani Zhang, Xingjian Shi, Junyuan Xie, Hao Ma, Irwin King, and Dit-Yan Yeung.
\newblock Gaan: Gated attention networks for learning on large and
  spatiotemporal graphs.
\newblock {\em arXiv preprint arXiv:1803.07294}, 2018.

\bibitem[\protect\citeauthoryear{Zhang \bgroup \em et al.\egroup
  }{2020}]{zhang2020adaptive}
Kai Zhang, Yaokang Zhu, Jun Wang, and Jie Zhang.
\newblock Adaptive structural fingerprints for graph attention networks.
\newblock In {\em International Conference on Learning Representations}, 2020.

\bibitem[\protect\citeauthoryear{Zhang \bgroup \em et al.\egroup
  }{2022}]{zhang2022graph}
Wentao Zhang, Ziqi Yin, Zeang Sheng, Yang Li, Wen Ouyang, Xiaosen Li, Yangyu
  Tao, Zhi Yang, and Bin Cui.
\newblock Graph attention multi-layer perceptron.
\newblock {\em arXiv preprint arXiv:2206.04355}, 2022.

\bibitem[\protect\citeauthoryear{Zhao \bgroup \em et al.\egroup
  }{2022}]{zhao2022learning}
Tong Zhao, Gang Liu, Daheng Wang, Wenhao Yu, and Meng Jiang.
\newblock Learning from counterfactual links for link prediction.
\newblock In {\em International Conference on Machine Learning}, pages
  26911--26926. PMLR, 2022.

\bibitem[\protect\citeauthoryear{Zhu \bgroup \em et al.\egroup
  }{2020}]{zhu2020beyond}
Jiong Zhu, Yujun Yan, Lingxiao Zhao, Mark Heimann, Leman Akoglu, and Danai
  Koutra.
\newblock Beyond homophily in graph neural networks: Current limitations and
  effective designs.
\newblock {\em Advances in Neural Information Processing Systems},
  33:7793--7804, 2020.

\end{thebibliography}
}

\end{document}